\documentclass[10pt, a4paper]{article}

\usepackage[final]{lrec2026} 

\title{Language Models as Semantic Augmenters for Sequential Recommenders}

\name{Mahsa Valizadeh, Xiangjue Dong, Rui Tuo, James Caverlee } 

\address{Texas A\&M University\\
        College Station, Texas, USA \\
         \{mvalizadeh, xj.dong, ruituo, caverlee\}@tamu.edu\\}

\abstract{
Large Language Models (LLMs) excel at capturing latent semantics and contextual relationships across diverse modalities. However, in modeling user behavior from sequential interaction data, performance often suffers when such semantic context is limited or absent. We introduce \textbf{LaMAR}, a LLM-driven semantic enrichment framework designed to enrich such sequences automatically. LaMAR leverages LLMs in a few-shot setting to generate auxiliary contextual signals by inferring latent semantic aspects of a user's intent and item relationships from existing metadata. These generated signals, such as inferred usage scenarios, item intents, or thematic summaries, augment the original sequences with greater contextual depth. We demonstrate the utility of this generated resource by integrating it into benchmark sequential modeling tasks, where it consistently improves performance. Further analysis shows that LLM-generated signals exhibit high semantic novelty and diversity, enhancing the representational capacity of the downstream models. This work represents a new data-centric paradigm where LLMs serve as intelligent context generators, contributing a new method for the semi-automatic creation of training data and language resources.
 \\ \newline , Sequential Behavior Modeling}

\begin{document}

\maketitleabstract

\section{Introduction}

Capturing users' dynamic preferences through their historical behaviors has been a central focus of sequential recommender systems~\citep{chen2018sequential,xie2022contrastive,liu2024mamba4rec,lin2024rella}. A persistent limitation is that these interaction histories are often represented as sparse sequences of item identifiers with minimal contextual information, making it difficult for models to infer nuanced user preferences, especially in cold-start scenarios or long-tail items. Meanwhile, large language models (LLMs) like ChatGPT and Gemini have demonstrated remarkable capabilities in capturing semantic relationships and generating coherent text based on minimal input. Their ability to reason about real-world concepts and infer latent intent makes them promising tools for enhancing recommendation systems with richer semantic context.

In this work, we propose \textbf{LaMAR} (Language Model-Augmented Recommendation), which leverages the reasoning capabilities of LLMs to enrich item representations through automatically generated semantic features. Instead of traditional feature engineering, LaMAR uses an LLM to generate auxiliary descriptions and latent signals that provide context to a user's interaction history. Concretely, given a user's past sequence of items, we prompt a large language model (with a carefully designed prompt and a few illustrative examples) to produce additional information, such as categorical tags or topics that summarize the user's interests, or an explanation of the possible intent or scenario underlying the sequence. These generated signals are then incorporated into the training of the sequential recommender (e.g., by encoding them into the model's input representations). By enriching each user's history with LLM-inferred context, the model gains a deeper understanding of preferences that would not be evident from the raw data alone. For instance, based on a user's last five product interactions, the LLM may infer an underlying intent (e.g., ``planning a camping trip'') or suggest a latent thematic category (e.g., ``outdoor adventure gear''). 

\begin{figure*}[t]
    \centering
    \includegraphics[width=0.98\textwidth]{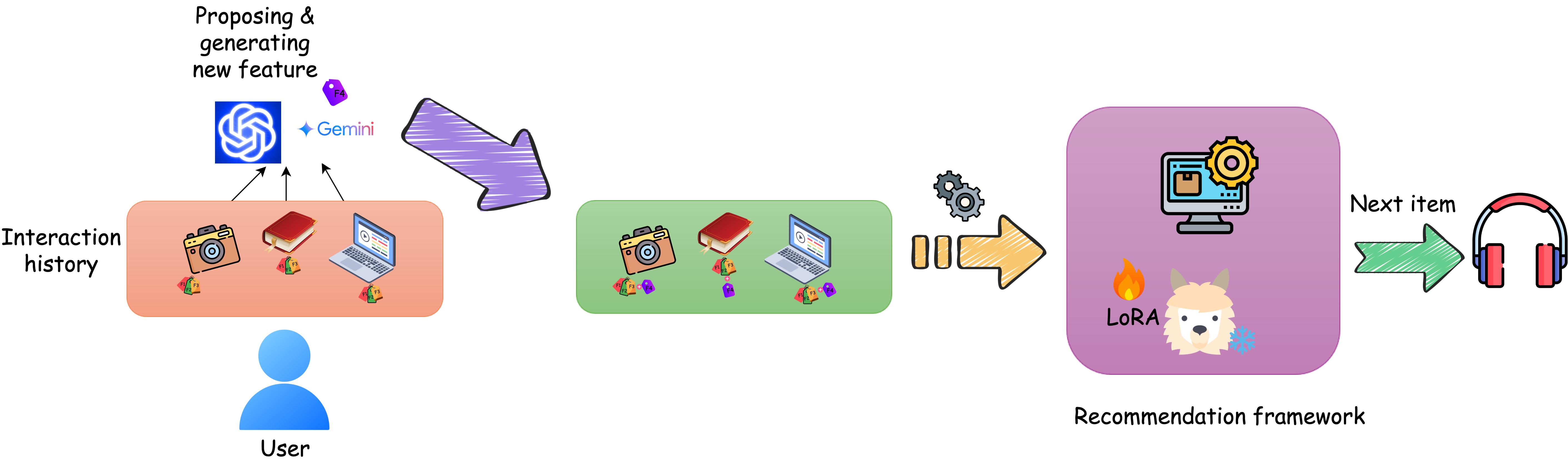}
    \caption{\textbf{Overview of the LaMAR framework.} It consists of two stages: (1) semantic signal generation using a prompted large language model, and (2) integration of the generated signals into a sequential recommendation model via full fine-tuning, or into the LLM through parameter-efficient LoRA tuning.}
    \label{fig:framework}
\end{figure*}

We conduct extensive experiments to evaluate the effectiveness of our framework on multiple public sequential recommendation datasets, comparing against state-of-the-art sequential recommendation baselines. Our results show that adding LLM-generated signals to user-item sequences leads to consistent performance gains across all key ranking metrics. To understand the contributions of the generated signals, we perform a thorough analysis of their diversity and uniqueness, and demonstrate their semantic richness. We find that the LLM is capable of contributing a wide range of complementary information rather than simply echoing a user's existing history, which expands the knowledge base of the recommender. Furthermore, we explore fine-tuning the LLM using the augmented sequences and observe additional improvements. Our code is here: \url{https://github.com/mahsavalizadeh/LaMAR}.

In summary, our contributions are as follows:

\begin{itemize}
    \item We introduce a semantic augmentation framework that uses LLMs to generate informative and semantically diverse contextual signals for sequential recommendation. These signals are designed to capture semantic relationships and enhance the richness of the dataset, providing additional context that sequential models can leverage.

    \item We conduct a deeper analysis to assess the extent of feature redundancy and representation diversity. Specifically, we investigate how similar or distinct the generated features are, identifying potential overlaps and unique contributions.
    
    \item We show that incorporating these signals into existing sequential recommendation models yields significant improvements across multiple datasets.

    \item We explore the impact of data augmentation on sequential recommendation tasks while fine-tuning LLMs on both the enriched and original datasets.  

\end{itemize}

\section{Related Work}
LLMs have recently transformed how we understand and represent semantics across a wide range of AI tasks, including recommendation. Their capacity for language understanding, content generation, and contextual reasoning has inspired many efforts to incorporate LLMs into recommender systems. 

Early studies primarily leveraged LLMs for semantic augmentation, enriching item or user representations and generating additional training data in recommender systems~\citep{lin2023can, liu2024large}.  For example, \citet{brinkmann2023product} use ChatGPT to extract attribute–value pairs from product descriptions, while KAR~\citep{xi2024towards} and LLM-Rec~\citep{lyu2023llm} leverage LLMs to produce auxiliary knowledge or enriched item descriptions. Other work explores broader content-enrichment strategies, including content summarization~\citep{liu2024once}, category descriptions~\citep{yada2024news}, taxonomy-guided augmentation~\citep{liang2024taxonomy}, factual property extraction~\citep{luo2024large}, and description-based augmentation for ID-based recommenders~\citep{ren2024enhancing}. Beyond feature enrichment, several approaches employ LLMs to produce synthetic interactions or simulate user behaviors
~\citep{wang2024large, huang2025large}, construct structured semantic representations like knowledge graphs~\citep{wang2024llmrg, liu2025understanding} or provide reasoning paths~\cite{bismay-etal-2025-reasoningrec}.

Before the emergence of LLM-based methods, architectures such as SASRec~\citep{kang2018self} and BERT4Rec~\citep{sun2019bert4rec} demonstrated strong performance through attention-based modeling of user-item interaction sequences. These and other methods -- e.g., \citep{chen2018sequential,xie2022contrastive} -- have typically modeled items simply by an ID, ignoring the rich semantic information contained in item content (e.g., text, images, or videos). Recent works have thus sought to bridge this gap by infusing item representations into sequential recommenders. Some explore \textit{semantic IDs} \citep{10.1145/3640457.3688190}, replacing simple item IDs with semantically-meaningful embeddings. Others leverage LLMs to directly model item content, e.g., RecFormer~\citep{li2023text} textualizes an item by flattening item characteristics (like title and description) into a language-based representation, while Rella~\citep{lin2024rella} applies zero-/few-shot LLM reasoning for recommendation.

In contrast, our work departs from these constrained enrichment methods. Rather than limiting the LLM to predefined augmentation tasks, we ask it to identify and generate unconstrained semantic signals—from nuanced gameplay experiences to product-specific usage contexts—that may not be present in the original metadata. Further, these signals are integrated directly into the recommendation model in a flexible, end-to-end fashion, instead of being used merely as auxiliary text features. This design enables the recommender to dynamically leverage diverse, new semantic cues, offering a complementary and more adaptive framework for improving recommendation quality, especially for items with limited interactions.
for improving recommendation quality, especially for items with limited interactions.

\begin{table}[t]
    \centering \small
    \resizebox{\linewidth}{!}{%
    \begin{tabular}{ll}
        \hline
        \textbf{Amazon Category} & \textbf{Generated Semantic Signals} \\
        \hline
        Industrial and Scientific & Primary Use Case \\
        Musical Instruments & Music Genre/Skill Suitability \\
        Arts, Crafts and Sewing & Primary Craft Purpose \\
        Office Products & Product Function/Use \\
        Video Games & Gameplay Experience Focus \\
        Pet Supplies & Pet's Specific Need \\
        \hline
    \end{tabular}}
        \caption{\textbf{LLM-generated semantic signals for different Amazon categories.} Each signal is inferred based on item metadata and tailored to capture domain-specific context, such as use case, purpose, or suitability.}
    \label{tab:proposed_features}
    \vspace{-10pt}
\end{table}

\section{Methodology of LaMAR}

Our proposed framework, LaMAR (Language Model-Augmented Recommendation), consists of two primary stages: (1) a semantic signal generation pipeline that uses LLMs to enrich user-item sequences with contextual cues, and (2) a signal integration stage that incorporates these signals into a sequential recommender system through fine-tuning to align LLM behavior with recommendation objectives.

\subsection{Semantic Signal Generation via LLMs}
\label{ssec:signal-generation}

In the first stage, we utilize the reasoning capabilities of LLMs to infer semantic signals that are not explicitly present in the dataset but can be derived from item metadata such as titles, brands, and categories. This new information captures latent contextual information about user intent and item semantics.
To generate contextually informative signals that enrich user-item sequences, we leverage the reasoning capabilities of LLMs through a prompting-based generation framework. Specifically, we adopt the ``Let's think step by step'' prompting strategy~\citep{kojima2022large}, which has proven effective in eliciting structured reasoning from language models. Our approach is aligned with the Automatic Chain-of-Thought (Auto-CoT) paradigm~\citep{zhang2022automatic}, wherein we first construct four examples using Zero-Shot-CoT, and then use them to guide the model in a few-shot setting to propose a new auxiliary signal.

Let $\mathcal{U}$ be the set of users, $\mathcal{I}$ the set of items. For a user $u \in \mathcal{U}$, their interaction sequence is $\mathbf{s}_u = [i_1, i_2, \dots, i_t]$, where $i_k \in \mathcal{I}$. Each item $i$ is associated with a set of structured attributes $\mathbf{x}_i = [\text{title}_i, \text{brand}_i, \text{category}_i, \cdots]$.
To infer a semantic signal $z_i$ for item $i$, we prompt a language model with its structured attributes:
\begin{equation}
z_i = \text{LLM}_\theta(\text{Prompt}(\mathbf{x}_i)),
\label{eq:signal_generation}
\end{equation}
where $\text{LLM}_\theta$ is a large language model with parameters $\theta$, and $\text{Prompt}(\cdot)$ denotes the few-shot prompting function that formats $\mathbf{x}_i$ into textual input. The output $z_i \in \mathcal{Z}$ is a natural language phrase or sentence that captures a latent aspect of the item, such as use case or thematic intent.

The LLM (e.g., GPT-3.5) outputs a new semantic signal tailored to the item’s profile and its probable user interaction context. These outputs are filtered for quality and stored as an additional item attribute to be later used during model training. The goal of this stage is to automatically derive features that encapsulate \textbf{implicit semantics}. For example, for the ``Pet Supplies'' domain, the model infers a ``Pet's Specific Need'' signal that highlights the intended style of use. For the ``Video Games'' domain, the model suggests ``Gameplay Experience Focus'' as an augmented signal. These inferred signals extend beyond raw metadata, adding interpretability and domain-relevant nuance.

\begin{table*}[t]
\centering \small
\begin{tabular}{p{2cm}rp{8cm}}
\hline
\textbf{Dataset} & \textbf{Category} & \textbf{Description} \\ 
\hline
\multirow{4}{*}{Office Products}  & Title & Ballet 2012 Square 12X12 Wall Calendar (Multilingual Edition) \\ &Brand & Browntrout Publishers (COR) \\ &Category& Office Products, Office \& School Supplies, Calendars, Planners \& Personal Organizers\\ &\textcolor{blue}{Product Function} &  \textcolor{blue}{Stylishly showcases beautiful ballet imagery while helping to organize dates and events throughout the year with its easy-to-read monthly grids.}\\  
\hline
\multirow{4}{*}{Musical Instruments} & Title & DW Drums 7000 Series Double Bass Drum Pedal\\ & Brand &Drum Workshop, Inc. \\ &Category & Musical Instruments Instrument Accessories Drum \& Percussion Accessories Drum Set Accessories Drum Accessories Bass Drum Pedals\\ & \textcolor{blue}{Music Genre} &\textcolor{blue}{Ideal for rock, metal, and jazz drumming styles, this pedal offers precision and responsiveness for both live performances and studio recordings, enhancing rhythmic expression.}\\  
\hline
\multirow{4}{*}{Video Games} & Title & America's Greatest Solitaire Games\\ & Brand & Wizardware Group \\ &Category & Video Games Mac Games\\ &\textcolor{blue}{Gameplay Experience Focus} & \textcolor{blue}{Engaging solitaire gameplay featuring a variety of classic card games, strategic challenges, and customizable settings, perfect for casual gamers on Mac.} \\  
\hline
\multirow{4}{*}{Industrial and Scientific } & Title & Arctic Silver 5 AS5-3.5G Thermal Paste\\ & Brand & Arctic Silver\\ &Category& Industrial \& Scientific Industrial Electrical Thermal Management Products Computer Heatsinks \\& \textcolor{blue}{Primary Use Case} & \textcolor{blue}{Thermal paste designed for computer heatsinks to improve heat transfer and cooling efficiency for optimal performance and longevity of electronic devices. }\\ 
\hline
\multirow{4}{*}{Arts, Crafts and Sewing } & Title &Electric Laser Guided Scissors Stainless Steel Blades \\& Brand & Salco\\ & Category & Arts, Crafts \& Sewing Crafting Craft Supplies Cutting Tools Scissors\\ & \textcolor{blue}{Primary Craft Purpose} & \textcolor{blue}{These laser-guided scissors provide precision cutting for various materials, enhancing crafting projects with clean and accurate results. Perfect for detailed and intricate designs.} \\  
\hline
\multirow{4}{*}{Pet Supplies } & Title & Hikari Usa Inc AHK01389 Staple 22lb, Medium \\& Brand & Hikari Usa Inc. \\ &Category& Pet Supplies Fish \& Aquatic Pets Food \\&\textcolor{blue}{Pet's Specific Need} &  \textcolor{blue}{A balanced and nutritious staple food to support the health and growth of medium-sized fish in aquatic environments.} \\  
\hline
\end{tabular}
\caption{Examples of LLM-generated semantic augmentations  (highlighted in \textcolor{blue}{blue}), including category and its specific value, for each Amazon dataset.}
\label{table:dataset_info}
\vspace{-10pt}
\end{table*}

\begin{table*}[t] 
\centering
\small
\resizebox{0.98\linewidth}{!}{%
\begin{tabular}{llcccccc}
\hline
\multirow{2}{*}{\textbf{Dataset}} & \multirow{2}{*}{\textbf{Metric}} & 
\multicolumn{1}{c@{\hspace{6pt}}}{\textbf{ID-Text}} & 
\multicolumn{5}{c}{\textbf{Text-based Methods}} \\
\cmidrule(lr{4pt}){3-3} \cmidrule(lr{4pt}){4-8}
 &  & \textbf{S$^3$Rec$^*$} & \textbf{ZESRec$^*$} & \textbf{UniSRec$^*$} &  \textbf{Recformer} &  \makecell{\textbf{LaMAR}\\\textbf{(GPT-4o)}} &
\makecell{\textbf{LaMAR}\\\textbf{(Gemini-1.5)}} \\
\hline
\multirow{6}{*}{Industrial and Scientific} 
 & N@10 &  0.0451 & 0.0843 & 0.0862 & 0.1052 & 0.1114 \textcolor{Green}{(+5.89\%)} & 0.1056 \textcolor{Green}{(+0.38\%)} \\
 & R@10 & 0.0804 & 0.1260 & 0.1255 & 0.1479 & 0.1524 \textcolor{Green}{(+3.04\%)} & 0.1504 \textcolor{Green}{(+1.69\%)} \\
 & N@50 &  -- & -- & -- & 0.1229 & 0.1286 \textcolor{Green}{(+4.64\%)} & 0.1238 \textcolor{Green}{(+0.73\%)} \\
 & R@50 &  -- & -- & -- & 0.2288 & 0.2313 \textcolor{Green}{(+1.09\%)} & 0.2338 \textcolor{Green}{(+2.19\%)} \\
 & MRR & 0.0392 & 0.0745 & 0.0786 & 0.0977 & 0.1044 \textcolor{Green}{(+6.86\%)} & 0.0976 \textcolor{Red}{(-0.10\%)} \\
 & AUC & -- & -- & -- & 0.7657 & 0.7658 \textcolor{Green}{(+0.01\%)} & 0.7723 \textcolor{Green}{(+0.86\%)}\\
\hline
\multirow{6}{*}{Musical Instruments}
 & N@10 &  0.0797 & 0.0694 & 0.0785 & 0.0814 & 0.0838 \textcolor{Green}{(+2.95\%)} & 0.0826 \textcolor{Green}{(+1.47\%)} \\
 & R@10 &  0.1110 & 0.1078 & 0.1119 & 0.1031 & 0.1076 \textcolor{Green}{(+4.36\%)} & 0.1064 \textcolor{Green}{(+3.20\%)} \\
 & N@50 & -- & -- & -- & 0.0952 & 0.0983 \textcolor{Green}{(+3.26\%)} & 0.0968 \textcolor{Green}{(+1.68\%)} \\
 & R@50 & -- & -- & -- & 0.1664 & 0.1747 \textcolor{Green}{(+4.99\%)} & 0.1722 \textcolor{Green}{(+3.49\%)} \\
 & MRR &  0.0755 & 0.0633 & 0.0740 & 0.0794 & 0.0813 \textcolor{Green}{(+2.39\%)} & 0.0800 \textcolor{Green}{(+0.75\%)} \\
 & AUC & -- & -- & -- & 0.7912 & 0.8058 \textcolor{Green}{(+1.84\%)} & 0.7972 \textcolor{Green}{(+0.76\%)} \\
\hline
\multirow{6}{*}{Arts, Crafts and Sewing} 
 & N@10 & 0.1026 & 0.0970 & 0.0894 & 0.1269 & 0.1282 \textcolor{Green}{(+1.02\%)} & 0.1283 \textcolor{Green}{(+1.10\%)} \\
 & R@10 & 0.1399 & 0.1349 & 0.1333 & 0.1579 & 0.1638 \textcolor{Green}{(+3.74\%)} & 0.1646 \textcolor{Green}{(+4.24\%)} \\
 & N@50 & -- & -- & -- & 0.1415 & 0.1443 \textcolor{Green}{(+1.98\%)} & 0.1447 \textcolor{Green}{(+2.26\%)} \\
 & R@50 & -- & -- & -- & 0.2255 & 0.2377 \textcolor{Green}{(+5.41\%)} & 0.2399 \textcolor{Green}{(+6.39\%)} \\
 & MRR & 0.1057 & 0.0870 & 0.0798 & 0.1218 & 0.1220 \textcolor{Green}{(+0.16\%)} & 0.1220 \textcolor{Green}{(+0.16\%)} \\
 & AUC & & -- & -- & 0.8275 & 0.8361 \textcolor{Green}{(+1.04\%)} & 0.8413 \textcolor{Green}{(+1.67\%)} \\
\hline
\multirow{6}{*}{Office Products} 
 & N@10  & 0.0911 & 0.0865 & 0.0919 & 0.1138 & 0.1163 \textcolor{Green}{(+2.20\%)} & 0.1173 \textcolor{Green}{(+3.07\%)} \\
 & R@10  & 0.1186 & 0.1199 & 0.1262 & 0.1408 & 0.1434 \textcolor{Green}{(+1.85\%)} & 0.1456 \textcolor{Green}{(+3.41\%)} \\
 & N@50 & -- & -- & -- & 0.1235 & 0.1262 \textcolor{Green}{(+2.17\%)} & 0.1274 \textcolor{Green}{(+3.16\%)} \\
 & R@50 & -- & -- & -- & 0.1854 & 0.1888 \textcolor{Green}{(+1.83\%)} & 0.1914 \textcolor{Green}{(+3.24\%)} \\
 & MRR & 0.0957 & 0.0797 & 0.0848 & 0.1084 & 0.1109 \textcolor{Green}{(+2.31\%)} & 0.1116 \textcolor{Green}{(+2.95\%)} \\
 & AUC & -- & -- & -- & 0.7586 & 0.7593 \textcolor{Green}{(+0.09\%)} & 0.7618 \textcolor{Green}{(+0.42\%)} \\
\hline
\multirow{6}{*}{Video Games} 
 & N@10 & 0.0532 & 0.0530 & 0.0580 & 0.0680 & 0.0715 \textcolor{Green}{(+5.15\%)} & 0.0710 \textcolor{Green}{(+4.41\%)} \\
 & R@10 & 0.0879 & 0.0844 & 0.0923 & 0.1039 & 0.1102 \textcolor{Green}{(+6.06\%)} & 0.1092 \textcolor{Green}{(+5.10\%)} \\
 & N@50 & -- & -- & -- & 0.0913 & 0.0950 \textcolor{Green}{(+4.05\%)} & 0.0948 \textcolor{Green}{(+3.83\%)} \\
 & R@50 & -- & -- & -- & 0.2120 & 0.2185 \textcolor{Green}{(+3.07\%)} & 0.2190 \textcolor{Green}{(+3.30\%)} \\
 & MRR & 0.0500 & 0.0505 & 0.0552 & 0.0643 & 0.0669 \textcolor{Green}{(+4.04\%)} & 0.0667 \textcolor{Green}{(+3.73\%)} \\
 & AUC & -- & -- & -- & 0.8912 & 0.8881 \textcolor{Red}{(-0.35\%)} & 0.8903 \textcolor{Red}{(-0.10\%)} \\
\hline
\multirow{6}{*}{Pet Supplies} 
 & N@10 &  0.0742 & 0.0754 & 0.0702 & 0.0968 & 0.0992 \textcolor{Green}{(+2.48\%)} & 0.0978 \textcolor{Green}{(+1.03\%)} \\
 & R@10 & 0.1039 & 0.1018 & 0.0933 & 0.1155 & 0.1213 \textcolor{Green}{(+5.02\%)} & 0.1212 \textcolor{Green}{(+4.94\%)} \\
 & N@50 & -- & -- & -- & 0.1049 & 0.1081 \textcolor{Green}{(+3.05\%)} & 0.1073 \textcolor{Green}{(+2.29\%)} \\
 & R@50 & -- & -- & -- & 0.1528 & 0.1620 \textcolor{Green}{(+6.02\%)} & 0.1648 \textcolor{Green}{(+7.85\%)} \\
 & MRR  & 0.0710 & 0.0706 & 0.0650 & 0.0936 & 0.0952 \textcolor{Green}{(+1.71\%)} & 0.0934 \textcolor{Red}{(-0.21\%)} \\
 & AUC & -- & -- & -- & 0.7888 & 0.7959 \textcolor{Green}{(+0.90\%)} & 0.7996 \textcolor{Green}{(+1.37\%)} \\
\hline
\end{tabular}}
\caption{Comparative analysis of LaMAR performance on semantic signals generated by \texttt{GPT-4o-mini} and \texttt{Gemini-1.5-flash} across multiple Amazon datasets. The text-based baselines refer to item representations using three features: Title, Brand, and Category. $^*$: results are reported from~\citet{li2023text}. N: NDCG; R: Recall.}
\label{results}
\vspace{-10pt}
\end{table*}

\subsection{Signal Integration and Model Alignment}

In the second stage, we use the generated semantic signals to enhance sequential recommendation through two mechanisms to improve representation capacity and downstream recommendation quality: (a) input-level integration into the user-item interaction history, and (b) alignment of the sequential recommendation models or LLM's generation behavior with the recommendation objective.

\smallskip
\noindent \textbf{Signal Integration.} LLM-generated signals are incorporated into the input representation of sequential recommendation models. 

We first augment the item representation to include the semantic signal:
\begin{equation}
\tilde{\mathbf{x}}_i = \mathbf{x}_i \cup \{ z_i \}.
\label{eq:augmented_item}
\end{equation}
As illustrated in Table~\ref{table:dataset_info}, we see an example for one specific product's ``Pet's Specific Need'' as \textit{A balanced and nutritious staple food to support the health and growth of medium-sized fish in aquatic environments.}

We then format each item as a text-based key-value pair and construct chronological user interaction sequences that include these enriched item representations. Thus, each user's enriched interaction sequence becomes:
\begin{equation}
\tilde{\mathbf{s}}_u = [\tilde{\mathbf{x}}_{i_1}, \tilde{\mathbf{x}}_{i_2}, \dots, \tilde{\mathbf{x}}_{i_t}]
\label{eq:augmented_sequence}
\end{equation}
This approach allows the model to consider \textbf{both explicit metadata and LLM-inferred semantics} when learning patterns from user history.

\smallskip
\noindent \textbf{Model Alignment.} Furthermore, the sequential recommenders or LLM can be fine-tuned using these augmented interaction sequences to better align its generative behavior with the needs of recommendation. 
To align the LM's generation behavior with recommendation utility, we fine-tune $\text{LM}_\theta$ on sequences enriched with $z_i$, training it to predict the next item description. The fine-tuning objective encourages the model to improve its understanding of useful semantics: 
\begin{equation}
\mathcal{L}_{\text{gen}}(\theta) = -\sum_{u \in \mathcal{U}} \log P_\theta(i_{t+1} \mid \tilde{\mathbf{s}}_u).
\label{eq:llm_finetuning}
\end{equation}
By training the model on sequences that include both structured metadata and generated semantic signals, the LLM learns to produce more relevant, context-aware outputs for future signal generation.

\begin{table*}[t]
\centering 
\small
\protect\resizebox{0.85\linewidth}{!}{%
\begin{tabular}{llllll}

\hline

\multirow{2}{*}{\textbf{Dataset}} & \multirow{2}{*}{\textbf{Metric}} 
& \multicolumn{2}{c}{\textbf{Llama}} 
& \multicolumn{2}{c}{\textbf{Qwen}} \\
\cmidrule(lr{4pt}){3-4} \cmidrule(lr{4pt}){5-6}
& & \textbf{Original} & \textbf{LaMAR (\%)} & \textbf{Original} & \textbf{LaMAR (\%)} \\
\midrule

\multirow{3}{*}{\makecell{Industrial and \\Scientific}} 
& R@10 & 0.7230 & 0.8810 (\textcolor{Green}{+21.85\%}) & 0.7680 & 0.8560 (\textcolor{Green}{+11.45\%}) \\
& N@10 & 0.6836 & 0.8773 (\textcolor{Green}{+28.48\%}) &  0.6996 & 0.8052 (\textcolor{Green}{+15.09\%}) \\
& MRR  & 0.6710 & 0.8761 (\textcolor{Green}{+30.57\%}) & 0.6777 & 0.7891 (\textcolor{Green}{+16.44\%}) \\
\hline

\multirow{3}{*}{\makecell{Arts, Crafts\\and Sewing}} 
& R@10 & 0.7350 & 0.8550 (\textcolor{Green}{+16.33\%}) & 0.8250 & 0.8530 (\textcolor{Green}{+3.39\%}) \\
& N@10 & 0.6947 & 0.8341 (\textcolor{Green}{+20.07\%}) & 0.7696 & 0.8047 (\textcolor{Green}{+4.56\%}) \\
& MRR  & 0.6818 & 0.8273 (\textcolor{Green}{+21.34\%}) & 0.7514 & 0.7893 (\textcolor{Green}{+5.05\%}) \\
\hline

\multirow{3}{*}{Video Games} 
& R@10 & 0.8080 & 0.9560 (\textcolor{Green}{+18.32\%}) & 0.9690 & 0.9750 (\textcolor{Green}{+0.62\%}) \\ 
& N@10 & 0.8056 & 0.9545 (\textcolor{Green}{+18.48\%}) & 0.9525 & 0.9666 (\textcolor{Green}{+1.48\%}) \\
& MRR  & 0.8048 & 0.9540 (\textcolor{Green}{+18.54\%}) & 0.9472 & 0.9639 (\textcolor{Green}{+1.76\%}) \\
\hline

\multirow{3}{*}{Pet Supplies} 
& R@10 & 0.9580 & 0.9660 (\textcolor{Green}{+0.83\%}) & 0.9180 & 0.9750 (\textcolor{Green}{+6.21\%}) \\ 
& N@10 & 0.9501 & 0.9608 (\textcolor{Green}{+1.13\%}) & 0.8878 & 0.9647 (\textcolor{Green}{+8.66\%}) \\
& MRR  & 0.9475 & 0.9591 (\textcolor{Green}{+1.22\%}) & 0.8780 & 0.9613 (\textcolor{Green}{+9.49\%}) \\
\hline

\multirow{3}{*}{\makecell{Musical\\Instruments}} 
& R@10 & 0.9370 & 0.9600 (\textcolor{Green}{+2.45\%}) & 0.9100 & 0.9410 (\textcolor{Green}{+3.41\%}) \\
& N@10 & 0.9294 & 0.9572 (\textcolor{Green}{+2.99\%}) & 0.8631 & 0.9149 (\textcolor{Green}{+6.00\%}) \\
& MRR  & 0.9269 & 0.9563 (\textcolor{Green}{+3.17\%}) & 0.8481 & 0.9064 (\textcolor{Green}{+6.88\%}) \\
\hline

\multirow{3}{*}{Office Products} 
& R@10 & 0.7210 & 0.8990 (\textcolor{Green}{+24.69\%}) & 0.6890 & 0.9160 (\textcolor{Green}{+32.98\%}) \\
& N@10 & 0.6912 & 0.8924 (\textcolor{Green}{+29.11\%}) & 0.6319 & 0.8943 (\textcolor{Green}{+41.54\%}) \\
& MRR  & 0.6816 & 0.8902 (\textcolor{Green}{+30.60\%}) & 0.6135 & 0.8873 (\textcolor{Green}{+44.63\%}) \\
\hline
\end{tabular}%
}
\caption{Comparative analysis of \texttt{LoRA} fine-tuned Llama and Qwen performance on the original (Title, Brand, and Category) versus LaMAR, which includes additional semantic signals generated by \texttt{GPT-4o-mini}. N: NDCG; R: Recall.}
\label{results-fine}
\vspace{-10pt}
\end{table*}

\section{Experiments}
In this section, we describe the experimental setup and aim to answer the following research question: Can integrating LLM-generated semantic signals (1) improve the performance of existing sequential recommenders and (2) enhance the recommendation capabilities of large language models?

\subsection{Setup}

\noindent \textbf{Models.} To assess the impact of the LLM-generated semantic signals, we employ RecFormer~\citep{li2023text}, a state-of-the-art framework for learning language representations in sequential recommendation. RecFormer models user behavior by processing a sequence of historical items as textual input and predicting the next item based on contextual understanding. It represents items as key-value attribute pairs and encodes their sequences using a bi-directional Transformer architecture. Inspired by Longformer, this model leverages specialized embeddings for item text to effectively capture sequential patterns. We also evaluate the performance of \texttt{Llama-3.2-3B-Instruct} model~\citep{touvron2023llama} and
\texttt{Qwen2.5-1.5B-Instruct} model~\citep{qwen2.5} with LoRA fine-tuning.

\smallskip
\noindent \textbf{Metrics.}
We use standard evaluation metrics, including NDCG@10~\citep{jarvelin2002cumulated}, Recall@10~\citep{schutze2008introduction}, NDCG@50, Recall@50, MRR~\citep{voorhees1999trec}, and AUC~\citep{hanley1982meaning}, and compare these values with those reported in \citep{li2023text}, which uses the RecFormer model without LLM-generated semantic signals. These metrics collectively measure both the ranking quality and retrieval accuracy of recommended items within the top positions.

\smallskip
\noindent \textbf{Implementation Details.}
To fine-tune \textit{Recformer} with generated features, we set batch size to $12$, learning rate to $2 \times 10^{-5}$, and weight decay to $0.01$. We set \textit{max\_item\_embeddings} $= 81$,  \textit{max\_token\_num} $= 256$, and \textit{max\_attr\_num} $= 4$. For fine-tuning with the original feature, we keep the default configuration of \texttt{Recformer} and only modify \textit{max\_token\_num } to $256$. Fine-tuning was conducted on $ 4 \times$ NVIDIA A5000 GPU (24GB memory).
We fine-tune the \texttt{Llama-3.2-3B-Instruct} model using a lightweight adapter-based approach, LoRA (Low-Rank Adaptation)~\citep{hu2022lora}, which introduces trainable low-rank decomposition matrices into the model's layers while keeping the original weight matrices frozen. The LoRA configuration included a rank of $r = 16$, $\alpha = 32$, and a dropout rate of $0.1$. The target modules for adaptation are \texttt{$q\_proj$}, \texttt{$k\_proj$}, and \texttt{$v\_proj$}. 
The learning rate is set to $2 \times 10^{-4}$ and fine-tuning was conducted on  $ 2 \times$ NVIDIA A5000 GPU (24GB memory). We configure \textit{per\_device\_train\_batch\_size} to $2$ and \textit{gradient\_accumulation\_steps} to $20$, resulting in an effective batch size of $80$. The same experimental configuration was applied to the \texttt{Qwen2.5-1.5B-Instruct} model, while using one GPU and a learning rate of $10^{-4}$.

\begin{table*}[t]
\centering
\small
\begin{tabular}{p{0.7cm}p{1cm}p{1cm}p{1cm}p{2.5cm}p{1.2cm}p{2.5cm}p{1.2cm}}
\hline
\textbf{Metric} &  \textbf{Original} &    \textbf{LaMAR} & \textbf{Improv.} &   \textbf{\makecell{LaMAR\\(prompt variant)}} & \textbf{Improv.} &  \textbf{\makecell{LaMAR\\(signal variant)}} & \textbf{Improv.} \\
\hline
N@10 &   0.1052 &  0.1114 & \textcolor{Green}{\textbf{+5.89\%}} & \centering 0.1080 & +2.66\% & \centering 0.1086 & +3.23\% \\
R@10 &  0.1479 &  0.1524 & \textcolor{Green}{\textbf{+3.04\%}} & \centering 0.1520 & +2.77\% & \centering 0.1477 & -0.14\% \\
N@50 &  0.1229 &  0.1286 & \textcolor{Green}{\textbf{+4.64\%}} & \centering 0.1255 & +2.12\% & \centering 0.1263 & +2.77\% \\
R@50 &  0.2288 &  0.2313 & +1.09\% & \centering 0.2324 & \textcolor{Green}{\textbf{+1.57\%}} & \centering 0.2290 & +0.09\% \\
MRR &  0.0977 &  0.1044 & \textcolor{Green}{\textbf{+6.86\%}}& \centering 0.1000 & +2.35\% & \centering 0.1022 & +4.61\% \\
AUC &  0.7657 &  0.7658 & \textcolor{Green}{\textbf{+0.01\%}} & \centering 0.7637 & -0.26\% & \centering 0.7624 & -0.43\% \\
\hline
\end{tabular}
\caption{\textbf{Comparison of LaMAR variants on the Scientific dataset.} We report performance using the original features (Title, Brand, Category), the standard LaMAR signal, and two variants: prompt-based (different prompting method) and signal-based (multiple signals). All semantic signals are generated using \texttt{GPT-4o-mini}.}
\label{metrics_comparison_sci}
  \vspace{-10pt}
\end{table*}

\subsection{Semantic Signal Generation}

In this work, we apply the signal generation framework across six categories from the Amazon review dataset~\cite{ni2019justifying}: Industrial and Scientific, Musical Instruments, Arts, Crafts and Sewing, Office Products, Video Games, and Pet Supplies. For each dataset, we apply the signal generation pipeline using \texttt{GPT-4o-mini} and \texttt{Gemini-1.5-flash-002} with a consistent 3-shot prompting configuration. The prompts used for generation are shown in Figure~\ref{propose_feature} and~\ref{generate_feature} in the Appendix. Table~\ref{tab:proposed_features} outlines the semantic signal generated for each domain, while Table~\ref{table:dataset_info} provides detailed description of the generated signals. Specifically, Table~\ref{tab:proposed_features} summarizes the nature of the semantic signal introduced per category, such as ``Primary Use Case'' for Industrial products or ``Gameplay Experience Focus'' for Video Games. Table~\ref{table:dataset_info} provides concrete instances where these signals are added to the item profile, showing how they complement the existing metadata. These enriched sequences are used to train and evaluate the RecFormer model.
These newly generated signals are integrated as a fourth attribute in addition to the standard triplet of Title, Brand, and Category. In our ablation studies, we also ablate the prompt and number of generated semantic signals in Section~\ref{ssec:ablation}. To evaluate the quality of LLM-generated semantic signals, we conduct a small-scale human study. Annotators rate 180 samples (15 per dataset per model) on two criteria (scored as 0, 0.5, or 1): (1) Relevancy: factual or contextual alignment with the input, and (2) Usefulness: added informative value beyond basic fields. The results (Table~\ref{quality}) indicate that the majority of generated signals are both relevant and informative, suggesting that hallucination is limited and the semantic signals are generally well-aligned with the task's objectives.

\begin{table}
\small
\centering
\begin{tabular}{lcc}
\hline
\textbf{Metric} & \textbf{GPT} & \textbf{Gemini} \\
\hline
Usefulness (Avg) & 13.67 & 13.29 \\
Usefulness (\%)  & \textcolor{Blue}{\textbf{91.13\%}} & \textcolor{Blue}{\textbf{88.60\%}} \\
Relevancy (Avg)  & 14.63 & 14.08 \\
Relevancy (\%)   & \textcolor{Blue}{\textbf{97.53\%}} & \textcolor{Blue}{\textbf{93.87\%}} \\
\hline
\end{tabular}
\caption{Results from the human evaluation study. Average scores and percentages for semantic signals generated by GPT and Gemini on ``Usefulness'' and ``Relevancy.'' Each dataset–model pair was rated on 15 samples (maximum total = 15; individual scores = 0, 0.5, 1).}
\label{quality}
\end{table}

\subsection{Can Semantic Signal Integration Enhance Sequential Recommenders?}

As the results show in Table~\ref{results}, we observe an improvement in performance across all datasets using both language models, highlighting the effectiveness of our proposed framework in enhancing sequential recommendation performance. 

For instance, for the Video Games dataset, the original model achieved an NDCG@10 of 0.0680 and a Recall@10 of 0.1039. After incorporating new signals through our framework, the model's performance improved: with GPT-generated signals, NDCG@10 increased to 0.0715 and Recall@10 to 0.1102, representing improvements of 5.15\% and 6.06\%, respectively; with Gemini-generated signals, NDCG@10 increased to 0.0710 and Recall@10 to 0.1092, corresponding to improvements of 4.41\% and 5.10\%, respectively. These results show that LaMAR enables the model to rank more relevant items higher and find more relevant items overall, improving performance regardless of whether GPT or Gemini is used.

\subsection{Can Semantic Signal Integration Improve LLM-Based Recommendation?}

To evaluate the effectiveness of data augmentation, we utilize two datasets: one with augmented data generated by GPT-4o-mini and another containing only the original data (where an item is represented with three features: Title, Brand, and Category), and compare their performance on a sequential recommendation task. For this purpose, we construct prompts consisting of five chronological user interactions, followed by a candidate pool of $20$ randomly selected items along with the ground truth item. We then prompt the model to recommend the next item from this pool. For our experiments, we adopt the \texttt{Llama-3.2-3B-Instruct} model~\citep{touvron2023llama} and \texttt{ Qwen2.5-1.5B-Instruct} model~\citep{qwen2.5} as the base LLMs and used $5{,}000$ prompts for fine-tuning. The results are presented in Table~\ref{results-fine}. The results confirm the effectiveness of our proposed framework; however, we observe that for datasets where the generated features are highly similar, the model shows only slight or no improvement. This indicates when using LLMs for recommendation systems, similar items can limit the benefit of data augmentation. Adding highly similar information can even lower the performance of the model, as it is shown in Table~\ref{results-fine} for ``Pet Supplies'' and ``Musical Instruments'' datasets, where the performance gains were minimal when similar features were added.

\section{Additional Results and Analysis}

\subsection{Ablation on Prompts and Signal Numbers.}
\label{ssec:ablation}

We further evaluate LaMAR's robustness on the Scientific dataset by examining two variants: an alternative prompting strategy and an expanded multi-signal setup. Specifically, we investigate whether modifying the prompt design or generating additional semantic signals improves recommendation performance. In the prompt variant, we modify the input format by including the dataset name, available features, and several random examples, prompting GPT-4o-mini to generate a new semantic signal. In the signal variant, we apply the standard LaMAR prompting strategy iteratively, generating and integrating a second signal.

As shown in Table~\ref{metrics_comparison_sci}, both variants improve over the original feature set (Title, Brand, Category). However, the standard LaMAR configuration yields the highest overall performance, with the largest gains in NDCG@10 (+5.89\%), Recall@10 (+3.04\%), and MRR (+6.86\%). The prompt variant shows slightly lower effectiveness, with performance drops of 2.66\%, 2.77\%, and 2.35\% in those same metrics, respectively, compared to LaMAR. The signal variant, while introducing a second semantic signal, does not lead to further gains and even slightly underperforms in Recall@10 and AUC, suggesting potential redundancy. These findings highlight that careful prompt design is critical, and that adding more semantic signals does not always translate to better recommendation performance.

\subsection{Signal Diversity Analysis}
To explore the novelty of the generated features for each dataset, we compare the semantic similarity, focusing on meaning and context rather than surface-level word overlap. We embed each sentence into a high-dimensional vector space, and compute the cosine similarity between pairs of vectors.  We use two embedding models: a general-purpose model, \texttt{all-MiniLM-L6-v2}, and a semantically focused model, \texttt{multi-qa-mpnet-base-cos-v1}. To identify similar texts, we apply five cosine similarity thresholds: $0.6$, $0.7$, $0.75$, $0.8$, and $0.9$. We categorize a text as ``highly similar'' in a given threshold when it is similar to more than $0.1\times$ the length of the dataset. 

\begin{figure}[!t]
  \centering

  \begin{subfigure}[t]{0.96\columnwidth} 
    \centering
    \includegraphics[width=\linewidth]{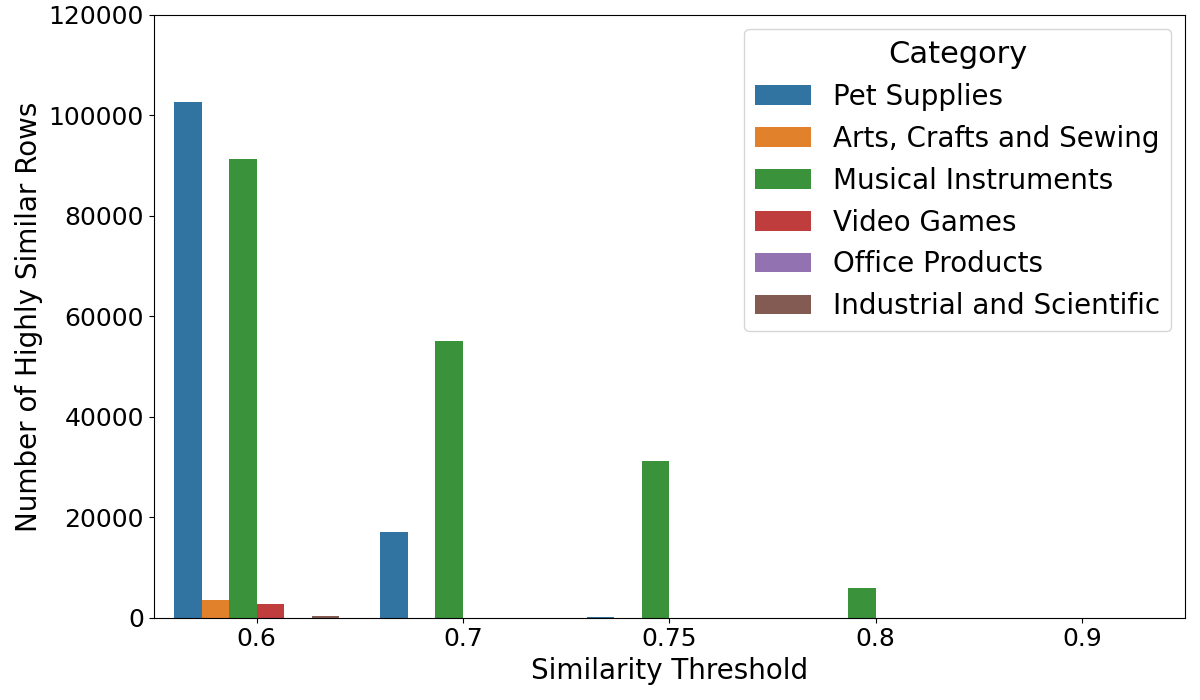}
    \caption{Semantic similarity analysis for \texttt{GPT-4o-mini}.}
    \label{fig:semantic_1}
  \end{subfigure}
  \hfill
  \begin{subfigure}[t]{0.96\columnwidth}
    \centering
    \includegraphics[width=\linewidth]{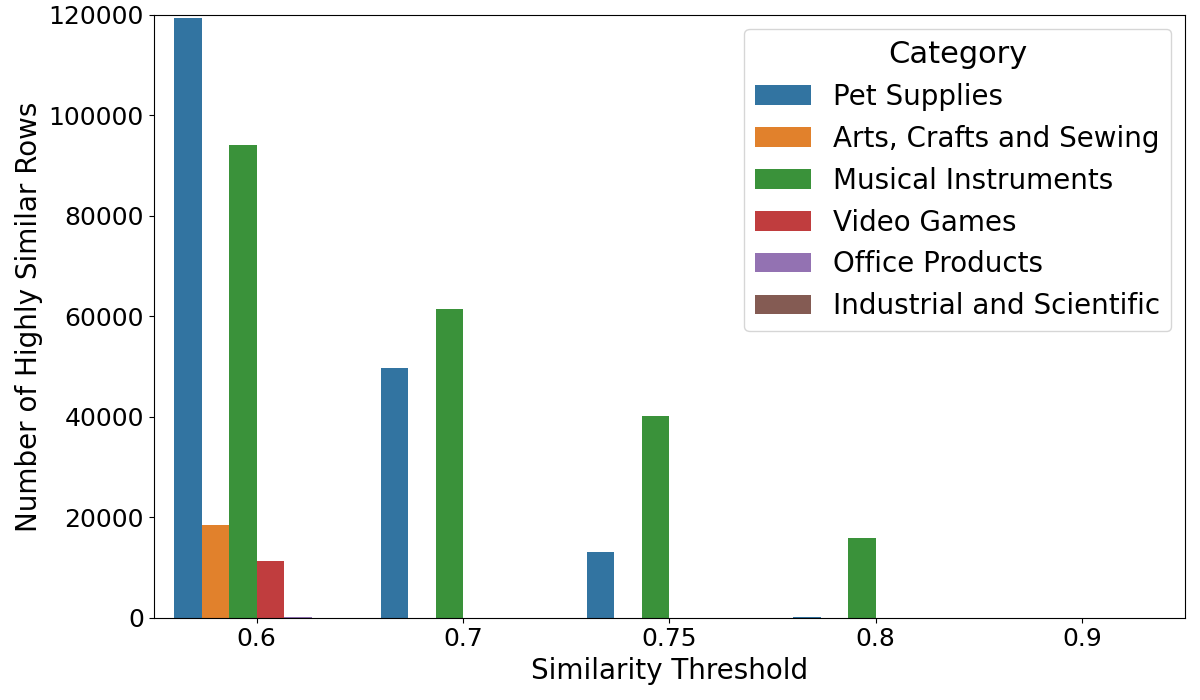}
    \caption{Semantic similarity analysis for \texttt{Gemini-1.5-Flash}.}
    \label{fig:semantic_2}
  \end{subfigure}

  \caption{Semantic similarity analysis across thresholds for signals generated by \texttt{GPT-4o-mini} and \texttt{Gemini-1.5-Flash}, using \texttt{multi-qa-mpnet-base-cos-v1}.}
  \label{fig:semantic similarity-comparison}
  \vspace{-10pt}
\end{figure}

\begin{figure}[!t]
  \centering

  \begin{subfigure}[t]{0.96\columnwidth}
    \centering
    \includegraphics[width=\linewidth]{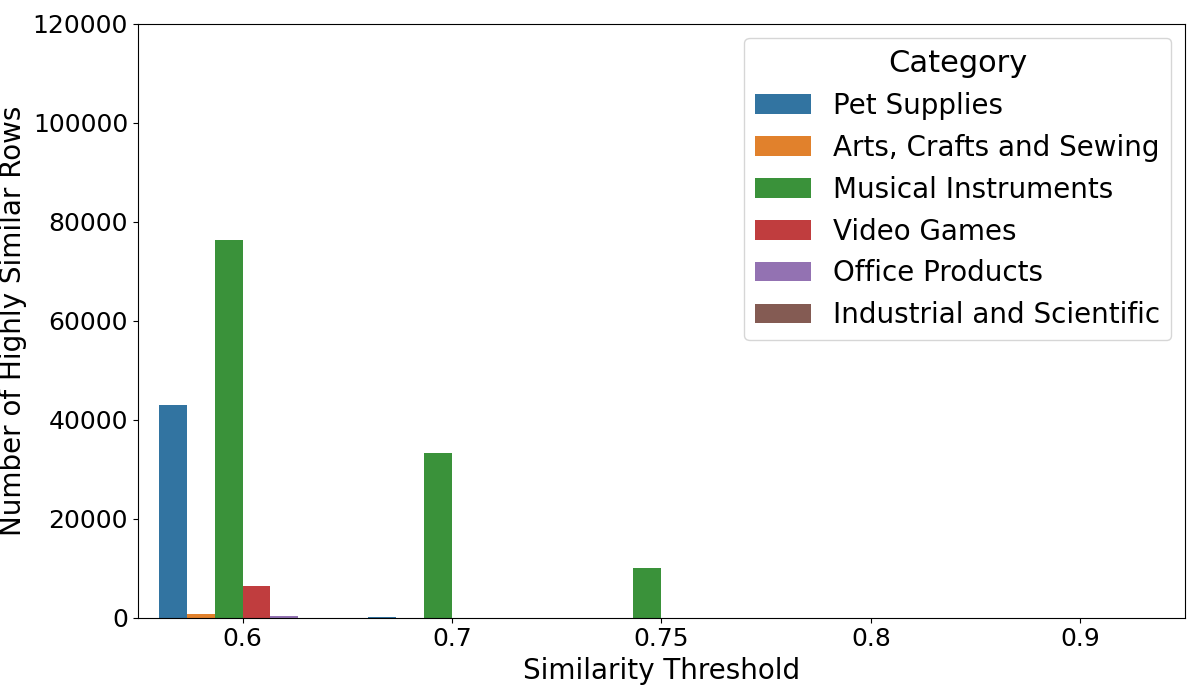}
    \caption{Semantic similarity analysis for \texttt{GPT-4o-mini}.}
    \label{fig:general_1}
  \end{subfigure}
  \hfill
  \begin{subfigure}[t]{0.96\columnwidth}
    \centering
    \includegraphics[width=\linewidth]{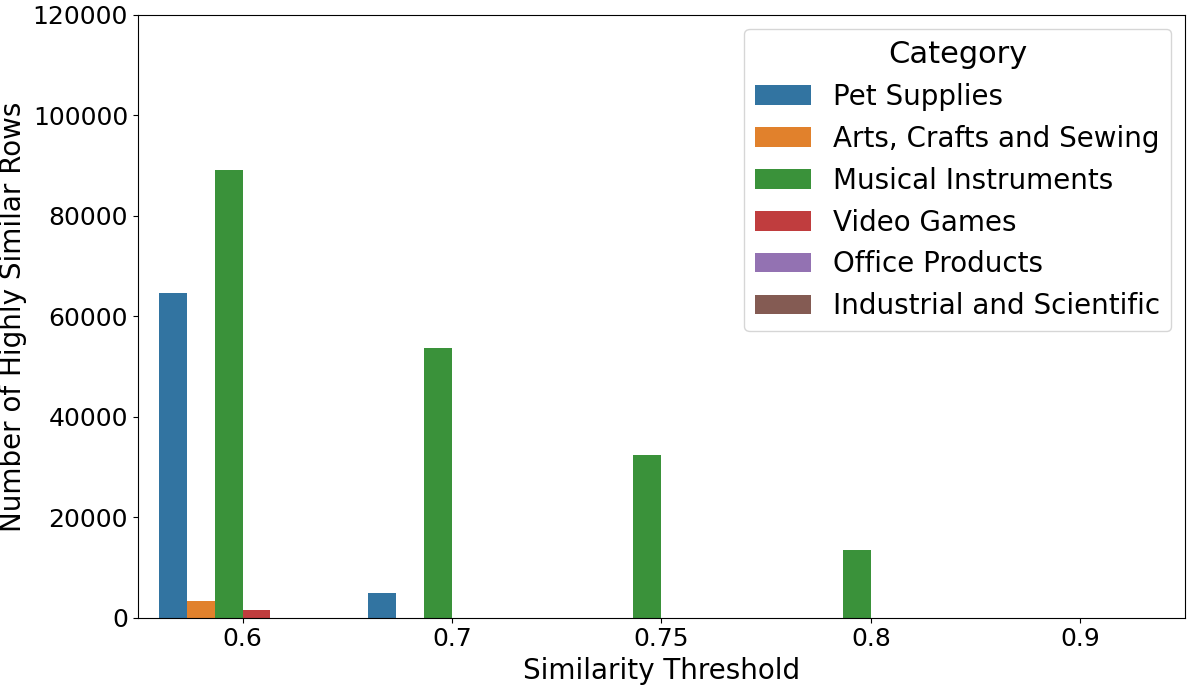}
    \caption{Semantic similarity analysis for \texttt{Gemini-1.5-Flash}.}
    \label{fig:general_2}
  \end{subfigure}

  \caption{Semantic similarity analysis across thresholds for signals generated by \texttt{GPT-4o-mini} and \texttt{Gemini-1.5-Flash}, using \texttt{all-MiniLM-L6-v2}.}
  \label{fig:general similarity-comparison}
  \vspace{-10pt}
\end{figure}

Figures~\ref{fig:semantic similarity-comparison} and ~\ref{fig:general similarity-comparison} represent the number of highly similar rows in each dataset for a feature generated by \texttt{GPT-4o-mini} and \texttt{Gemini-1.5-Flash}. The plot reveals that for features generated by both models, the semantic model detects a high number of similar rows at lower thresholds ($0.6$ -- $0.7$) especially for categories like Pet Supplies and Musical Instruments, indicating that the LLM may have reused similar phrasing or templates in these domains -- for example, recurring patterns involving common pet types or limited music genres. 

This suggests that during generation, certain categories are more prone to semantic repetition.
In addition, features generated by \texttt{Gemini-1.5-Flash} show higher similarity compared to those generated by \texttt{GPT-4o-mini} when comparing general and semantic similarity. Importantly, when we look beyond the $0.7$ threshold and exclude the Pet Supplies and Musical Instruments categories, the number of highly similar rows drops dramatically. This means that in most other domains, each generated feature is semantically distinct from over $90\%$ of the dataset, demonstrating strong content diversity. This provides confidence that LLMs can generate varied and original outputs in less templated categories.

\subsection{Generalization to a Different Domain}
To assess the generalizability of our approach beyond Amazon product categories, we conducted additional experiments on the MovieLens-1M dataset. In this setting, we used the movie title and genre as base features. Since the dataset does not include storyline information, we enriched it by incorporating storyline data obtained through web scraping. Furthermore, we introduced a new semantic signal, ``how the movie ends'',  generated by prompting \texttt{GPT-4o-mini} via LaMAR to enrich item representations. Table \ref{tab:recformer_movielens} summarizes the results, showing consistent improvements across multiple metrics. These results demonstrate that LaMAR generalizes effectively across domains, and the proposed semantic signal contributes meaningful improvements even outside the e-commerce setting.

\begin{table}
\small
\centering
\begin{tabular}{lccc}
\hline
\textbf{Metric} & \textbf{Recformer} & \makecell{\textbf{LaMAR}\\\textbf{(GPT-4o)}} & \makecell{\textbf{Improv.}\\\textbf{(\%)}}\\
\hline
NDCG@10     & 0.0384 & 0.0453 & \textcolor{Green}{+17.97\%} \\
Recall@10   & 0.0774 & 0.0926 & \textcolor{Green}{+19.64\%} \\
NDCG@50     & 0.0710 & 0.0782 & \textcolor{Green}{+10.14\%} \\
Recall@50   & 0.2288 & 0.2444 & \textcolor{Green}{+6.82\%}  \\
MRR         & 0.0367 & 0.0409 & \textcolor{Green}{+11.44\%} \\
AUC         & 0.8264 & 0.8241 & \textcolor{Red}{-0.27\%} \\

\hline
\end{tabular}
\caption{Comparing Recformer and LaMAR.}
\vspace{-10pt}
\label{tab:recformer_movielens}
\end{table}

\section{Conclusion}

We propose LaMAR, a framework that uses LLMs to generate semantic signals for augmenting sequential recommendation via few-shot prompting, eliminating the need for manual feature engineering. Experiments across domains show consistent performance gains, demonstrating the effectiveness of LLMs as scalable, data-centric semantic augmenters.

\section{Bibliographical References}\label{sec:reference}

\bibliographystyle{lrec2026-natbib}
\bibliography{lrec2026-example}

\clearpage
\newpage
\appendix

\section{Prompts}
\label{sec:appendix}
\begin{figure}[ht]
    \centering
    \includegraphics[width=.99\linewidth]{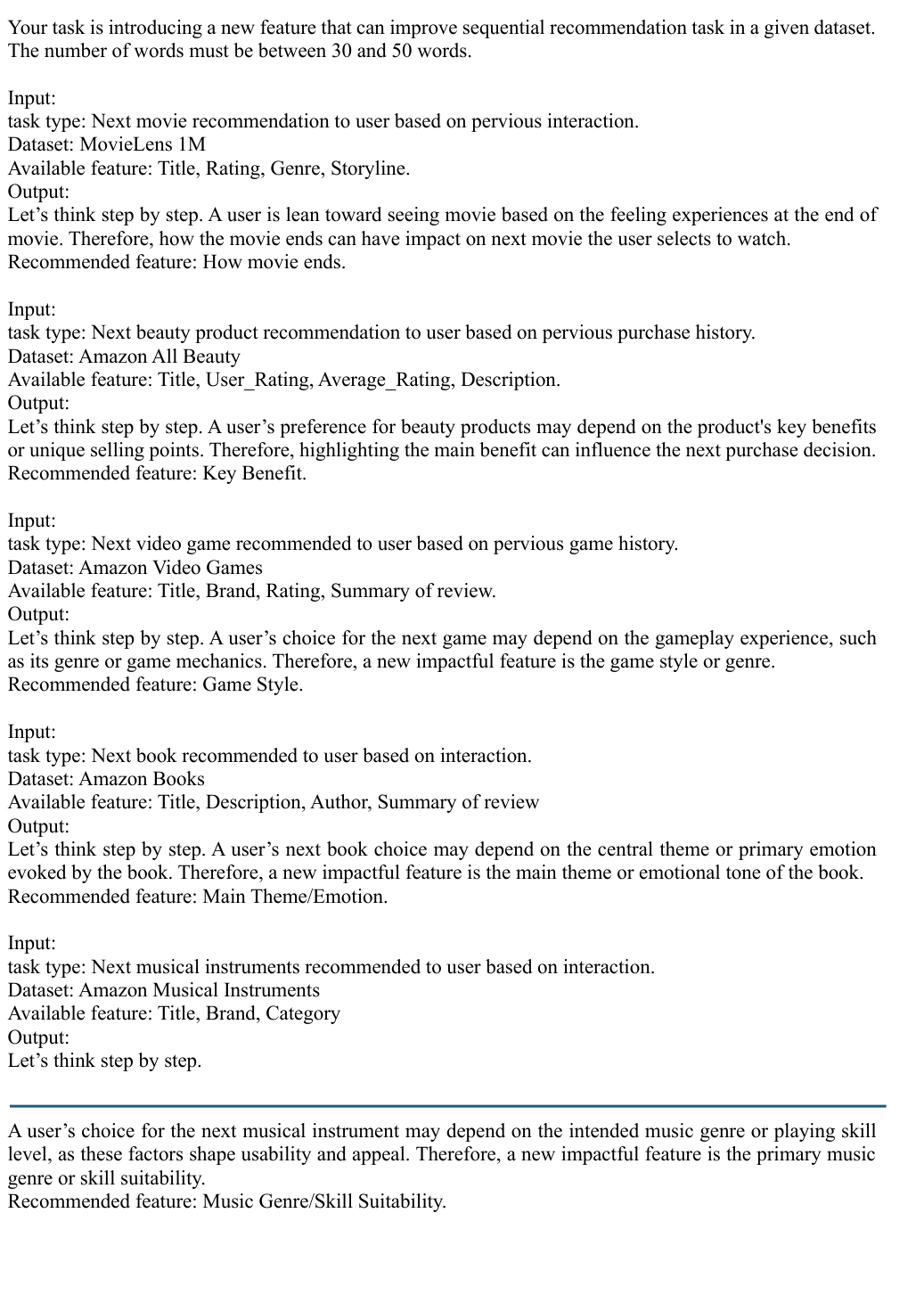}
    \caption{An example of prompt used for proposing a new relevant signal.}
    \label{propose_feature}
\end{figure}

\begin{figure}[ht]
    \centering
    \includegraphics[width=.99\linewidth]{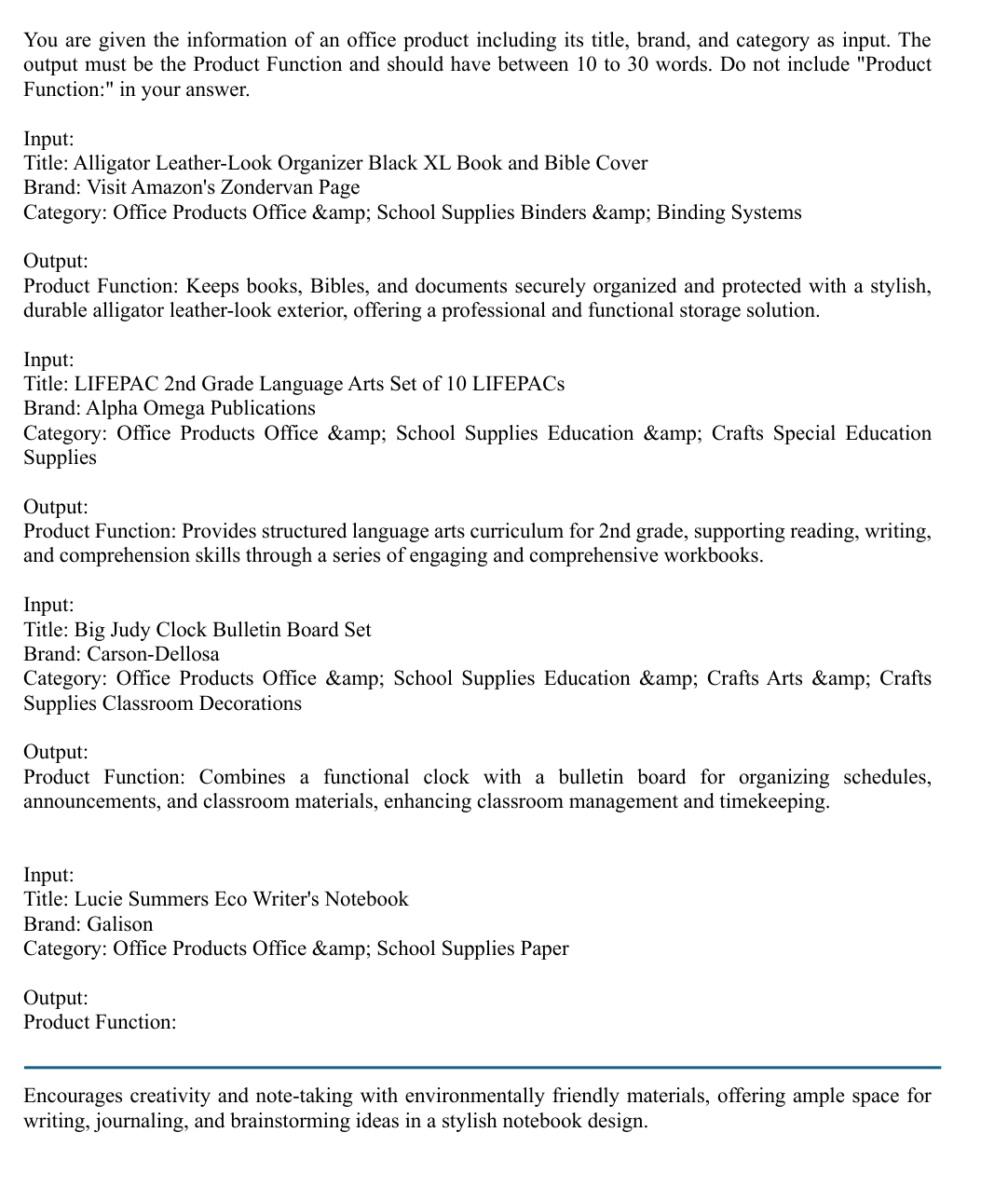}
    \caption{An example of prompt used for generating the proposed signal.}
    \label{generate_feature}
\end{figure}

\begin{figure}[!ht]
    \centering
    \begin{subfigure}[b]{0.99\linewidth}
        \centering
        \includegraphics[width=\linewidth]{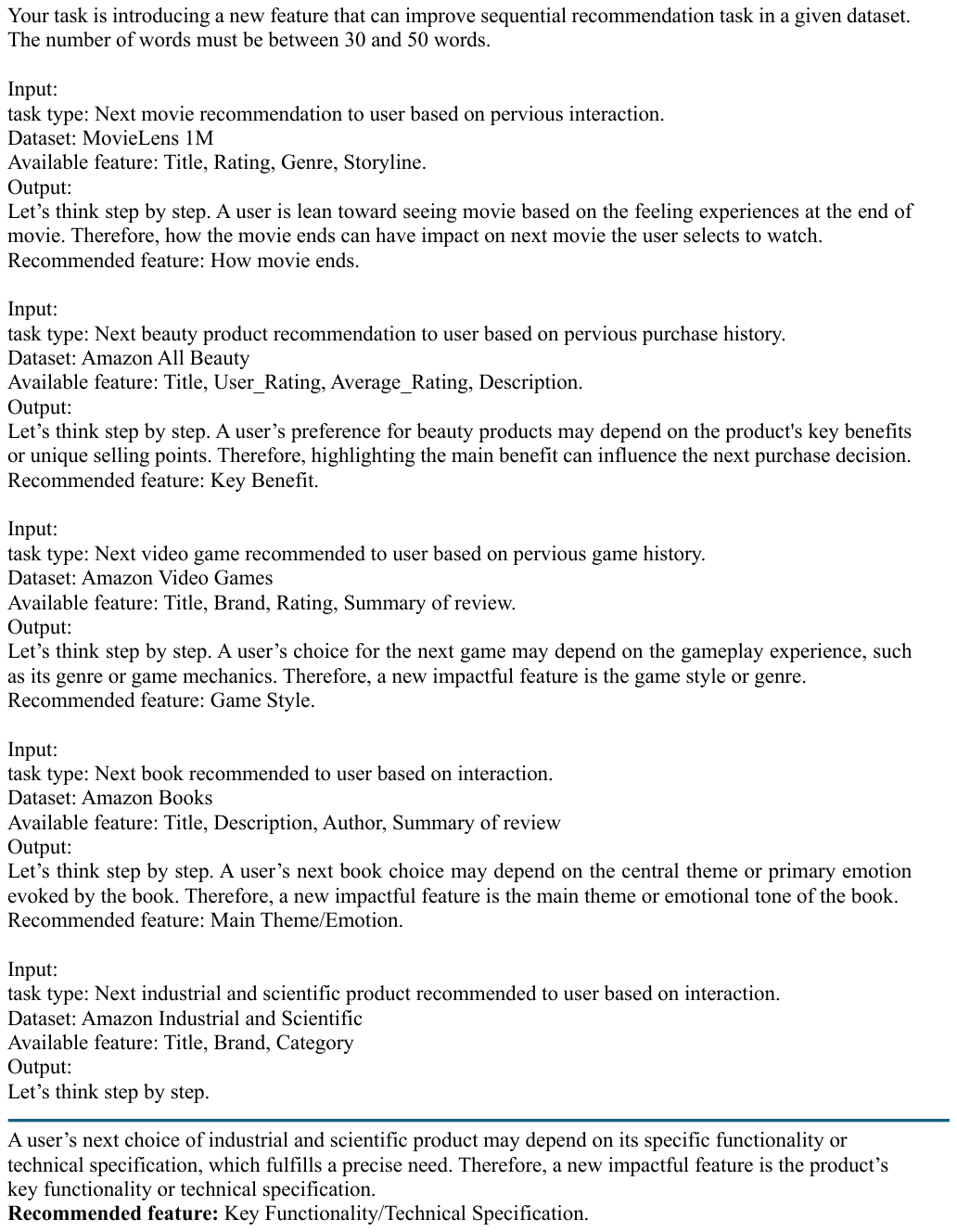}
        \caption{Example prompt used for proposing the fifth semantic signal -- first part.}
        \label{generate_feature_5_1}
    \end{subfigure}
    
    \vspace{1em}

    \begin{subfigure}[b]{0.99\linewidth}
        \centering
        \includegraphics[width=\linewidth]{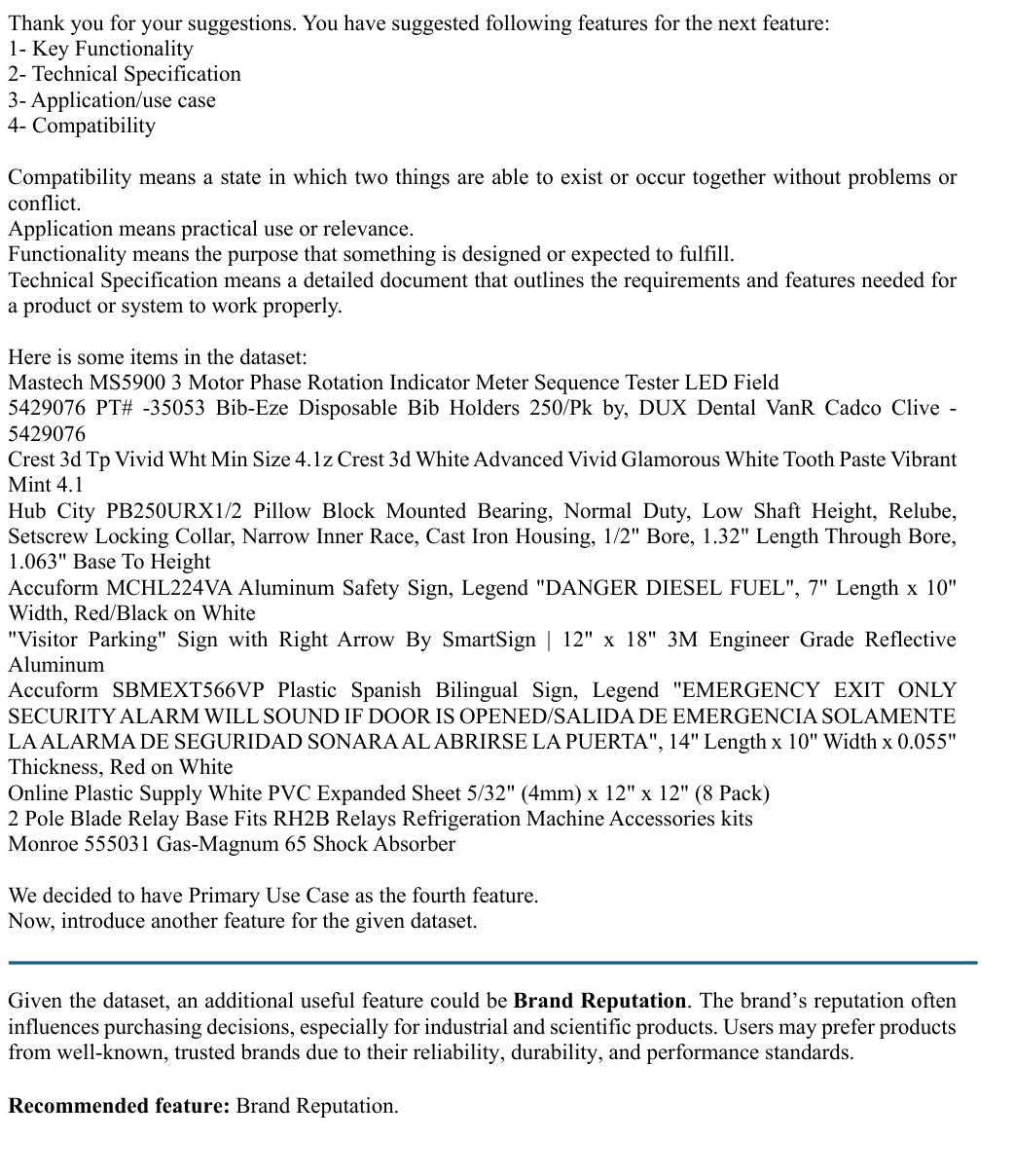}
        \caption{Example prompt used for proposing the fifth semantic signal -- second part.}
        \label{generate_feature_5_2}
    \end{subfigure}
    
    \caption{Examples of prompts used for proposing the fifth semantic signal.}
    \label{generate_feature_5_combined}
\end{figure}

An example of prompt employed for proposing new semantic signal and for generating the detailed description of LLM-generated semantic signals are shown in Figures~\ref{propose_feature} and \ref{generate_feature} respectfully. The prompt used to generate the fifth feature described in \ref{ssec:ablation} is also shown in \ref{generate_feature_5_combined}.

\begin{figure}[ht]
    \centering
    \includegraphics[width=.99\linewidth]{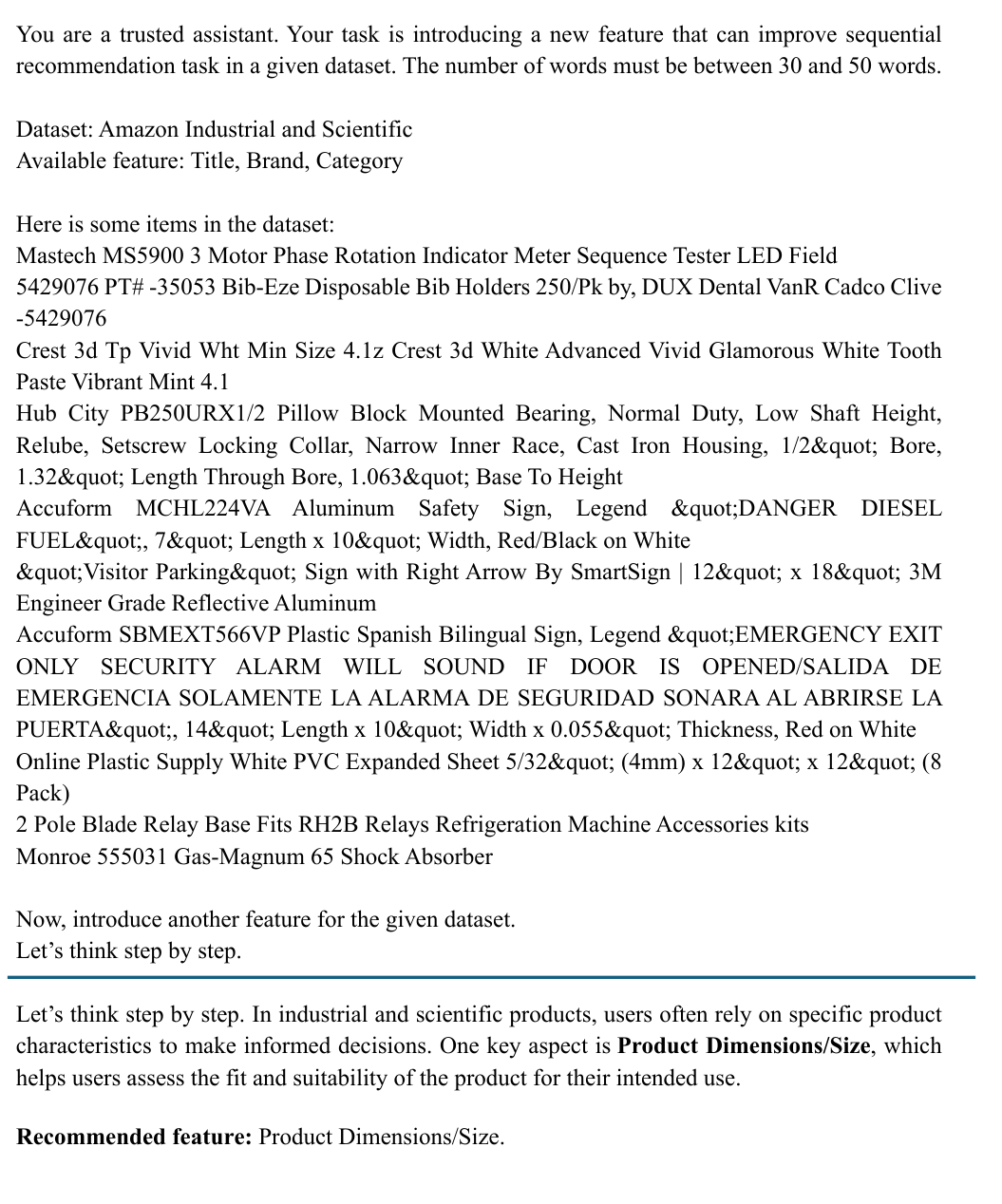}
    \caption{An example of the prompt used for generating through prompt variant.}
    \label{generate_pv}
\end{figure}

\section{More Implementation Details}
\subsection{Metrics}
We evaluate the effectiveness of our framework using standard metrics commonly adopted in recommendation research, including NDCG@10~\cite{jarvelin2002cumulated}, Recall@10~\cite{schutze2008introduction}, NDCG@50, Recall@50, MRR~\cite{voorhees1999trec}, and AUC~\cite{hanley1982meaning, he2017translation}. These metrics collectively measure both the ranking quality and retrieval accuracy of recommended items within the top positions. NDCG (Normalized Discounted Cumulative Gain) assesses the position-weighted relevance of recommended items, while Recall@K evaluates the proportion of relevant items retrieved among the top-K. MRR (Mean Reciprocal Rank) emphasizes the rank of the first correct item, and AUC (Area Under the ROC Curve) captures the overall discriminative ability of the model across all thresholds. We report these values for all experimental configurations and directly compare them to the baseline results reported in~\cite{li2023text}, which uses the RecFormer model without LLM-generated features. This comparison allows us to quantify the contribution of our semantic augmentation framework across diverse datasets and evaluation dimensions.

\subsection{Baselines}
We employ three types of methods as our baseline similar to~\cite{li2023text}, including approaches that use item IDs as primary inputs while incorporating item text as auxiliary information ($S^3-Rec$); and approaches have exclusively item texts as input (ZESRec, UniSRec, Recformer).
$S^3-Rec$~\cite{zhou2020s3} improves sequential recommendation by leveraging data correlations for self-supervision signals generation and data representations improvement. ZESRec~\cite{ding2021zero} is trained on an old dataset and generalizes to a new one with no overlapping users or items. UniSRec~\cite{hou2022towards} utilizes a lightweight MoE-based module to integrate textual item representations.

\subsection{Models}
Table~\ref{models} shows the models and their licenses.
\begin{table}[t]
\centering
\resizebox{\linewidth}{!}{%
\begin{tabular}{lp{5cm}l}
\hline
\textbf{Model} &  \textbf{Link} &    \textbf{License} \\ \hline
Llama-3.2-3B-Instruct & \url{https://huggingface.co/meta-llama/Llama-3.2-3B-Instruct} & llama3.2\\
GPT-4o-mini & \url{https://platform.openai.com/docs/models/gpt-4o-mini} & OpenAI \\
Gemini-1.5-flash & \url{https://cloud.google.com/vertex-ai/generative-ai/docs/models/gemini/1-5-flash}& Google \\
Qwen2.5-1.5B-Instruct & \url{https://huggingface.co/Qwen/Qwen2.5-1.5B-Instruct} & Apache License 2.0 \\
\hline
\end{tabular}}
\caption{Models.}
\label{models}
\end{table}

\subsection{Data Statistics}
Table~\ref{data-statistics} shows the statistics of the dataset.

\begin{table}
\small
\resizebox{0.9\linewidth}{!}{%
\begin{tabular}{lrrrrrr}
\hline
\textbf{Category} & \textbf{Users} & \textbf{Items} & \textbf{Interactions}  \\
\hline
Industrial and Scientific      & 11,041 & 5,327  & 76,896  \\
Musical Instruments     & 27,530 & 10,611 & 231,312  \\
Arts, Crafts and Sewing  & 56,210 & 22,855 & 492,492  \\
Office Products          &101,501 & 27,932 & 798,914 \\
Video Games           & 11,036 & 15,402 & 100,255 \\
Pet Supplies             & 47,569 & 37,970 & 420,662 \\
\hline
\end{tabular}}
\caption{Data statistics.}
\label{data-statistics}
  \vspace{-10pt}
\end{table}

\subsection{Data Preprocessing}
For RecFormer, we follow the data processing procedure described in \cite{li2023text}. Specifically, we use the datasets provided by the original data sources and filter out items with missing titles. User interactions are grouped by user and sorted in ascending order by timestamp to form sequential input. For each item, we extract key attributes—such as title, categories, and brand—and represent them as key-value pairs. For fine-tuning LLaMA, each training example is constructed from five chronologically ordered items from a user's interaction history, enriched with their corresponding semantic features, with the next item in the sequence used as the ground-truth target.

\end{document}